\documentclass[10pt,onecolumn,twoside]{article}

\usepackage{epsfig}
\usepackage{graphicx}
\usepackage{amsmath}
\usepackage{amsfonts}
\usepackage{amssymb}
\usepackage{bm}
\usepackage{placeins}
\usepackage{cite}
\usepackage{fullpage}
\usepackage{url}

\def\defn{\,\triangleq\,}
\def\etal{\emph{et al.}}

\def\betabf{\bm{\beta}}
\def\ybf{\mathbf{y}}
\def\ubf{\mathbf{u}}
\def\vbf{\mathbf{v}}
\def\xbf{\mathbf{x}}
\def\ebf{\mathbf{e}}
\def\psibf{\bm{\psi}}
\def\phibf{\bm{\phi}}
\def\sigmabf{\bm{\sigma}}
\def\Hbf{\mathbf{H}}
\def\Bbf{\mathbf{B}}
\def\Rbf{\mathbf{R}}
\def\Ibf{\mathbf{I}}

\def\sbf{\bm{s}}
\def\zbf{\mathbf{z}}

\def\phibfhat{\widehat{\bm{\phi}}}

\def\phibftilde{\tilde{\bm{\phi}}}

\def\R{\mathbb{R}}

\def\Rcal{\mathcal{R}}
\def\Lcal{\mathcal{L}}
\def\Gcal{\mathcal{G}}
\def\Tcal{\mathcal{T}}

\def\argmin{\mathop{\mathrm{arg\,min}}}
\def\prox{\mathrm{prox}}
\def\diag{\mathrm{diag}}
\def\rank{\mathrm{rank}}

\begin{document}


\title{Depth Superresolution using Motion Adaptive Regularization}


\author{Ulugbek~S.~Kamilov%
\thanks{U.~S.~Kamilov (email: kamilov@merl.com) and P.~T.~Boufounos (email: petrosb@merl.com)
are with Mitsubishi Electric Research Laboratories, 201 Broadway, Cambridge,
MA 02140, USA.}
\hspace{0.05em} and Petros~T.~Boufounos}

\maketitle


\begin{abstract}
Spatial resolution of depth sensors is often significantly lower compared to that of conventional optical cameras. Recent work has explored the idea of improving the resolution of depth using higher resolution intensity as a side information. In this paper, we demonstrate that further incorporating temporal information in videos can significantly improve the results. In particular, we propose a novel approach that improves depth resolution, exploiting the space-time redundancy in the depth and intensity using motion-adaptive low-rank regularization. Experiments confirm that the proposed approach substantially improves the quality of the estimated high-resolution depth. Our approach can be a first component in systems using vision techniques that rely on high resolution depth information.
\end{abstract}


\section{Introduction}
\label{Sec:Introduction}
One of the important challenges in computer vision applications is
obtaining high resolution depth maps of observed scenes. A number of
common tasks, such as object reconstruction, robotic navigation, and
automotive driver assistance can be significantly improved by
complementing intensity information from optical cameras with high
resolution depth maps. However, with current sensor technology, direct
acquisition of high-resolution depth maps is very expensive. 

The cost and limited availability of such sensors imposes significant
constraints on the capabilities of vision systems and has dampened the
adoption of methods that rely on high-resolution depth maps. Thus, the
literature has flourished with methods that provide numerical
alternatives to boost the spatial resolution of the measured depth
data.

\begin{figure}[t]
\begin{center}
\includegraphics[width=8cm]{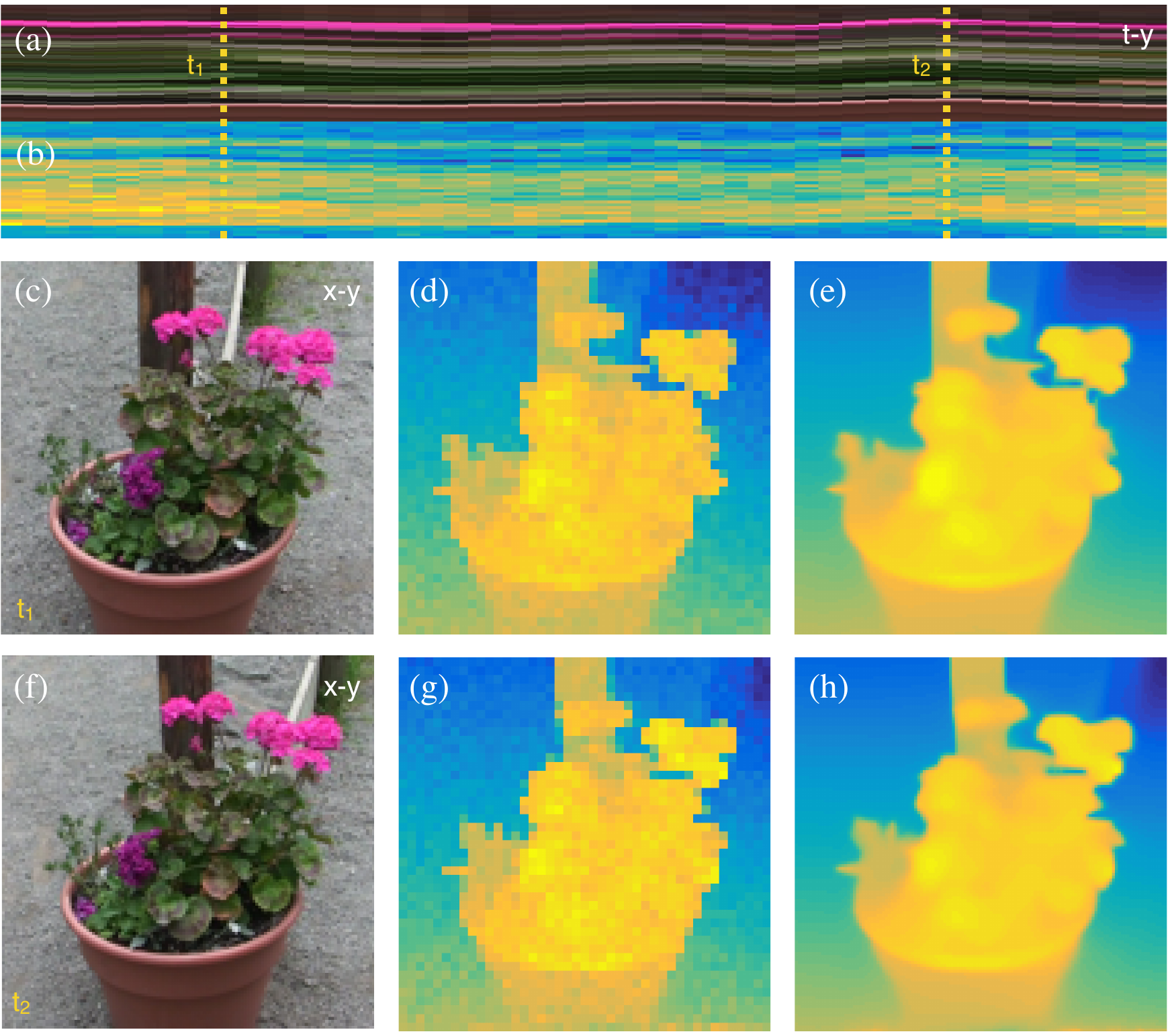}
\end{center}
\caption{Our motion adaptive method recovers a high-resolution depth
  sequence from high-resolution intensity and low-resolution depth
  sequences by imposing rank constraints on the depth patches: (a) and
  (b) $t$-$y$ slices of the color and depth sequences, respectively,
  at a fixed $x$; (c)--(e) $x$-$y$ slices at $t_1 = 10$; (f)--(h)
  $x$-$y$ slices at $t_2=40$; (c) and (f) input color images; (d) and
  (g) input low-resolution and noisy depth images; (e) and (h)
  estimated depth images.}
\label{Fig:MainFigure}
\end{figure}

One of the most popular and widely investigated class of techniques
for improving the spatial resolution of depth is guided depth
superresolution. These techniques jointly acquire the scene using a
low-resolution depth sensor and a high-resolution optical camera. The
information acquired from the camera is subsequently used to
superresolve the low-resolution depth map. These techniques exploit
the property that both modalities share common features, such as edges
and joint texture changes. Thus, such features in the optical camera
data provide information and guidance that significantly enhances the
superresolved depth map.

To-date, most of these methods operate on a single snapshot of the
optical image and the low-resolution depth map. However, most
practical uses of such systems acquire a video from the optical camera
and a sequence of snapshots of the depth map. The key insight in our
paper is that information about one particular frame is replicated, in
some form, in nearby frames. Thus, frames across time can be exploited
to superresolve the depth map and significantly improve such
methods. The challenge is finding this information in the presence of
scene, camera, and object motion between
frames. Figure~\ref{Fig:MainFigure} provides an example, illustrating
the similarity of images and depth maps across frames.

A key challenge in incorporating time into depth estimation is that
depth images change significantly between frames. This results in
abrupt variations in pixel values along the temporal dimension and may
lead to significant degradation in the quality of the result. Thus, it
is important to compensate for motion during estimation. To that end,
the method we propose exploits space-time similarities in the data
using motion adaptive regularization. Specifically, we identify and
group similar depth patches, which we superresolve and regularize
using a rank penalty.

Our method builds upon prior work on patch-based methods and low-rank
regularization, which were successfully applied to a variety of
practical estimation problems.  It further exploits the availability
of optical images which provide a very robust guide to identify and
group similar patches, even if the depth map has very low
resolution. Thus, the output of our iterative algorithms is robust to
operating conditions. Our key contributions are summarized as follows:
\begin{itemize}

\item We provide a new formulation for guided depth superresolution,
  incorporating temporal information. In this formulation, the high
  resolution depth is determined by solving an inverse problem that
  minimizes a cost. This cost includes a quadratic data-fidelity term,
  as well as a new motion adaptive regularizer based on a low-rank
  penalty on groups of similar patches.

\item We develop two optimization strategies for solving our
  estimation problem. The first approach is based on exact
  optimization of the cost via alternating direction method of
  multipliers (ADMM). The second approach uses a simplified algorithm
  that alternates between enforcing data-consistency and low-rank
  penalty.

\item We validate our approach experimentally and demonstrate it
  delivers substantial improvements. In particular, we compare
  several algorithmic solutions to the problem and demonstrate that:
  (a) availability of temporal information significantly improves the
  quality of estimated depth; (b) motion adaptive regularization is
  crucial for avoiding artifacts along temporal dimension; (c) using
  intensity during block matching is essential for optimal performance.

\end{itemize}


\section{Related Work}
\label{Sec:RelatedWork}

In the last decade, guided depth superresolution has received
significant attention. Early work by Diebel and
Thrun~\cite{Diebel.Thrun2005} showed the potential of the approach by
modeling the co-occurence of edges in depth and intensity with Markov
Random Fields (MRF). Kopf \etal~\cite{Kopf.etal2007} and Yang
\etal~\cite{Yang.etal2007} have independently proposed an alternative
approach based on joint bilaterial filtering, where intensity is used
to set the weights of the filter. The bilaterial filtering approach
was further refined by Chan \etal~\cite{Chan.etal2008} who
incorporated the local statistics of the depth and by Liu
\etal~\cite{Liu.etal2013} who used geodesic distances for determining
the weights. Dolson \etal~\cite{Dolson.etal2010} extended the approach
to dynamic sequences to compensate for different data rates in the
depth and intensity sensors. He \etal~\cite{He.etal2010, He.etal2013}
proposed a guided image filtering approach, improving edge
preservation.

More recently, sparsity-promoting regularization---an essential
component of compressive sensing~\cite{Candes.etal2006,
  Donoho2006}---has provided more dramatic improvements in the quality
of depth superresolution. For example, Li \etal~\cite{Li.etal2012}
demonstrated improvements by combining dictionary learning and sparse
coding algorithms. Ferstl \etal~\cite{Ferstl.etal2013} relied on
weighted total generalized variation (TGV) regularization for imposing
a piecewise polynomial structure on depth. Gong
\etal~\cite{Gong.etal2014} combined the conventional MRF approach with
an additional term promoting transform domain sparsity of the depth in
an analysis form. In their recent work, Huang
\etal~\cite{Huang.etal2015} use the MRF model to jointly segment the
objects and recover a higher quality depth. Schuon
\etal~\cite{Schouon.etal2009} performed depth superresolution by
taking several snapshots of a static scene from slightly displaced
viewpoints and merging the measurements using sparsity of the weighted
gradient of the depth.

Many natural images contain repetitions of similar patterns and
textures. Current state-the-art image denoising methods such as
nonlocal means (NLM)~\cite{Buades.etal2010} and block matching and 3D
filtering (BM3D)~\cite{Dabov.etal2007} take advantage of this
redundancy by processing the image as a structured collection of
patches. The original formulation of NLM was extended by Yang and
Jacob~\cite{Yang.Jacob2013} to more general inverse problems via
introduction of specific NLM regularizers. Similarly, Danielyan
\etal~\cite{Danielyan.etal2012} have proposed a variational approach
for general BM3D-based image reconstruction that inspired the current
work. In the context of guided depth superresolution, NLM was used by
Huhle \etal~\cite{Huhle.etal2010} and Park \etal~\cite{Park.etal2011}
for reducing the amount of noise in the estimated depth. Lu
\etal~\cite{Lu.etal2014} combined a block-matching procedure with
low-rank constraints for enhancing the resolution of a single depth
image.

Our paper extends prior work on depth supperresolution by introducing a new variational
formulation that imposes low-rank constraints in the
regularization. Furthermore, our formulation is motion-adaptive,
resulting in substantial improvement of the quality of the estimated
depth.


\section{Proposed Approach}
\label{Sec:ProposedApproach}

Our approach estimates the high-resolution depth map by minimizing a
cost function that---as typical in such problems---combines a
data-fidelity term and a regularizer. Specifically, we impose a
quadratic data fidelity term that controls the error between the
measured and estimated depth values. The regularizer groups similar
depth patches from multiple frames and penalizes the rank of the
resulting structure. Since we use patches from multiple frames, our
method is implicitly motion adaptive. Thus, by effectively combining
multiple views of the scene, it yields improved depth estimates.

\begin{figure}[t]
\begin{center}
\includegraphics[width=8cm]{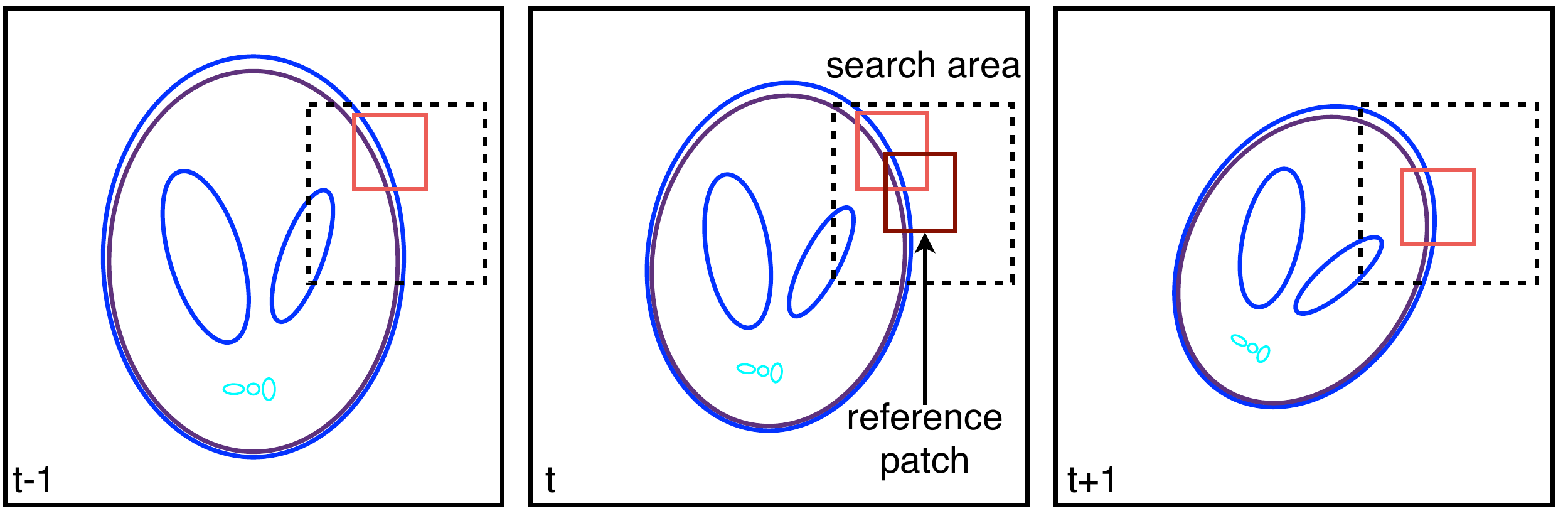}
\end{center}
\caption{Illustration of the block matching within a space-time search
  area. The area in the current frame $t$ is centered at the reference
  patch. Search is also conducted in the same window position in
  multiple temporally adjacent frames. Similar patches are grouped
  together to construct a block $\betabf_p = \Bbf_p \phibf$.}
\label{Fig:BlocMatching}
\end{figure}

\subsection{Problem Formulation}
\label{Sec:ProblemFormulation}

The depth sensing system collects a set of measurements denoted
$\{\psibf_t\}_{t \in [1 \dots T]}$. Each measurement is considered as
a downsampled version of a higher resolution depth map $\phibf_{t} \in
\R^N$ using a subsampling operator $\Hbf_t$. Our end goal is to
recover this high-resolution depth map $\phibf_{t}$ for all $t$.

In the remainder of this work, we use $N$ to denote the number of
pixels in each frame, $T$ to denote the number of temporal frames, and
$M$ to denote the total number of depth measurements. Furthermore,
$\psibf \in \R^M$ denotes the vector of all the measurements, $\phibf
\in \R^{NT}$ the complete sequence of high-resolution depth maps, and
$\Hbf \in \R^{M \times NT}$ the complete subsampling operator.  We
also have available the sequence of high-resolution intensity images
from the optical camera, denoted $\xbf \in \R^{NT}$.

Using the above, a forward model for the depth recovery problem is
given by
\begin{equation}
\psibf = \Hbf \phibf + \ebf,
\end{equation}
where $\ebf \in \R^M$ denotes the measurement noise. Thus, our
objective becomes to recover high-resolution depth given the measured
data $\psibf$ and $\xbf$, and the sampling operator $\Hbf$.

As typical in such problems, we formulate the depth estimation task as
an optimization problem
\begin{equation}
\label{Eq:OptimizationProblem}
\phibfhat = \argmin_{\phibf \in \R^{NT}} \left\{\frac{1}{2} \|\psibf -
\Hbf\phibf\|_{\ell_2}^2 + \sum_{p = 1}^P \Rcal(\Bbf_p\phibf)\right\},
\end{equation}
where $\frac{1}{2} \|\psibf - \Hbf\phibf\|_{\ell_2}^2$ enforces data
fidelity and $\sum_{p = 1}^P \Rcal(\Bbf_p\phibf)$ is a
regularization term that imposes prior knowledge about the depth map.

We form the regularization term by constructing sets of patches from
the image. Specifically, we first define an operator $\Bbf_p$, for
each set of patches $p \in [1, \dots, P]$, where $P$ is the number of
such sets constructed.  The operator extracts $L$ patches of size $B$
pixels from the depth image frames in $\phibf$. As illustrated in
Fig.~\ref{Fig:BlocMatching}, each block $\betabf_p = \Bbf_p\phibf \in
\R^{B \times L}$ is obtained by first selecting a reference patch and
then finding $L-1$ similar patches within the current frame as well as
the adjacent temporal frames.

To determine similarity and to group similar patches together we use
the intensity image as a guide. To reduce the computational complexity
of the search, we restrict it to a space-time window of fixed size
around the reference patch. We perform the same block matching
procedure for the whole space-time image by moving the reference patch
and by considering overlapping patches in each frame. Thus, each pixel
in the signal $\phibf$ may contribute to multiple blocks.

The adjoint $\Bbf^T_p$ of $\Bbf_p$ simply corresponds to placing the
patches in the block back to their original locations in $\phibf$. It
satisfies the following property
\begin{equation}
\sum_{p = 1}^P \Bbf^T_p \Bbf_p = \Rbf,
\end{equation}
where $\Rbf = \diag(r_1, \dots, r_N) \in \R^{NT \times NT}$ and $r_n$
denotes the total number of references to the $n$th pixel by the
matrices $\{\Bbf_p\}_{p = 1,\dots,P}$. Therefore, the depth image
$\phibf$ can be expressed in terms of an overcomplete representation
using the blocks
\begin{equation}
\phibf = \Rbf^{-1} \sum_{p = 1}^P \Bbf_p^T \Bbf_p \phibf.
\end{equation}

\subsection{Rank regularization}
\label{Sec:RankRegularization}
Each block, represented as a matrix, contains multiple similar
patches, i.e., similar columns. Thus, we expect the matrix to have a low rank,
making rank a natural regularizer for the problem
\begin{equation}
\label{Eq:RankRegularizer}
\Rcal(\betabf) = \rank(\betabf).\quad(\betabf \in \R^{B \times L})
\end{equation}

By seeking a low-rank solution to~\eqref{Eq:OptimizationProblem}, we
exploit the similarity of blocks to guide superresolution while
enforcing consistency with the observed data. However, the rank
regularizer~\eqref{Eq:RankRegularizer} is of little practical interest
since its direct optimization is intractable. The most popular
approach around this, first proposed by Fazel
in~\cite{fazel2002matrix}, is to convexify the rank by replacing it
with the nuclear norm:
\begin{equation}
\label{Eq:NuclearNormRegularizer}
\Rcal(\betabf) = \lambda\|\betabf\|_\ast \defn \lambda \hspace{-1em}\sum_{k = 1}^{\min(B, L)} \sigma_k(\betabf),
\end{equation}
where $\sigma_k(\betabf)$ denotes the $k$th largest singular value of
$\betabf$ and $\lambda > 0$ is a parameter controlling the amount of
regularization. 

In addition to its convexity, the nuclear norm is an appealing penalty
to optimize because it also has a closed form proximal operator:
\begin{align}
\prox_{\lambda \|\cdot\|_\ast}&(\psibf) \defn \argmin_{\betabf \in \R^{B \times L}}\left\{\frac{1}{2}\|\psibf - \betabf\|_F^2 + \lambda \|\betabf\|_\ast\right\} \nonumber \\
&= \ubf \, \eta_\lambda (\sigmabf(\psibf)) \, \vbf^T,
\end{align}
where $\psibf = \ubf \sigmabf \vbf^T$ is the singular value
decomposition (SVD) of $\psibf$ and $\eta_\lambda$ is the
soft-thresholding function applied to the diagonal matrix $\sigmabf$.

\begin{figure}[t]
\begin{center}
\includegraphics[width=8cm]{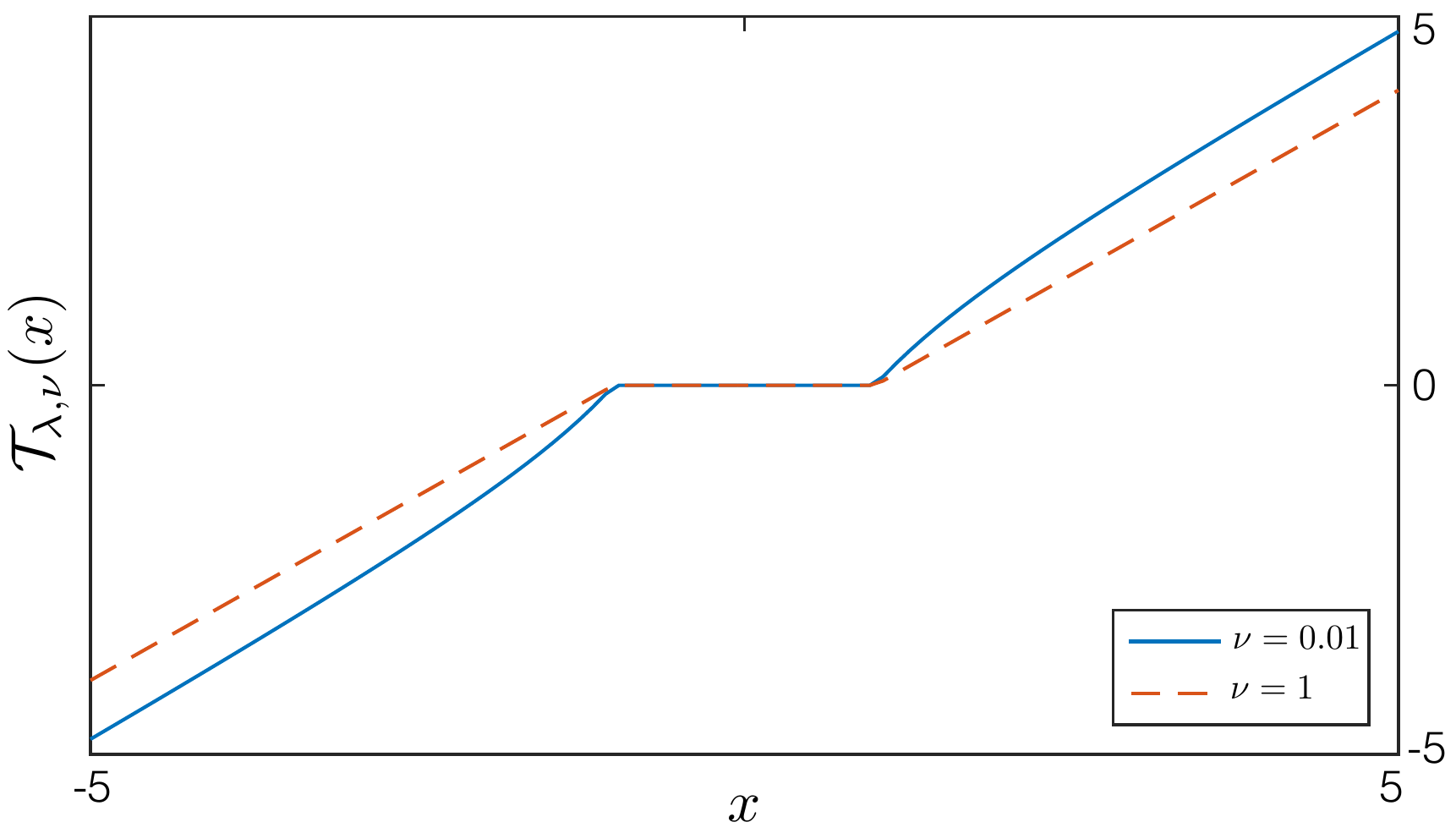}
\end{center}
\caption{Illustration of the $\nu$-shrinkage operator
  $\Tcal_{\nu,\lambda}$ for a fixed $\lambda = 1$ at two values of
  $\nu$. For $\nu = 1$ the shrinkage is equivalent to
  soft-thresholding, while for $\nu \rightarrow 0$ it approaches
  hard-thresholding}
\label{Fig:Shrink}
\end{figure}

Recent work has shown that nonconvex regularizers consistently
outperform nuclear norm by providing stronger denoising capability
without losing important signal compoenents~\cite{Chartrand2012,
  Hu.etal2012, Yoon.etal2014}. In this paper, we use the nonconvex
generalization to the nuclear norm proposed by
Chartrand~\cite{Chartrand2012}
\begin{equation}
\label{Eq:GeneralHuberRegularizer}
\Rcal(\betabf) = \lambda \Gcal_{\lambda, \nu}(\betabf) \defn \lambda \hspace{-1em}\sum_{k = 1}^{\min(B, L)} g_{\lambda, \nu}(\sigma_k(\betabf)),
\end{equation}
Here, the scalar function $g_{\lambda, \nu}$ is designed to satisfy
\begin{equation}
\min_{x \in \R} \left\{\frac{1}{2}|x-y|^2 + \lambda g_{\lambda, \nu}(x)\right\} =  h_{\lambda, \nu}(x),
\end{equation}
where $h_{\lambda, \nu}$ is the $\nu$-Huber function
\begin{equation}
h_{\lambda, \nu}(x) \defn
\begin{cases} 
   \frac{|x|^2}{2\lambda} & \text{if } |x| < \lambda^{1/(2-\nu)} \\
   \frac{|x|^\nu}{\nu}-\delta       & \text{if } |x| \geq \lambda^{1/(2-\nu)},
  \end{cases}
\end{equation}
with $\delta \defn (1/\nu - 1/2)\lambda^{\nu/(2-\nu)}$.
Although $g_{\lambda, \nu}$ is nonconvex and has no closed form formula, its proximal operator does admit a closed form expression
\begin{align}
\label{Eq:HuberProximal}
\prox_{\lambda \Gcal_{\lambda,\nu}}&(\psibf) \defn \argmin_{\betabf \in \R^{B \times L}}\left\{\frac{1}{2}\|\psibf - \betabf\|_F^2 + \lambda \Gcal_{\lambda, \nu}(\betabf)\right\} \nonumber \\
&= \ubf \, \Tcal_{\lambda, \nu} (\sigmabf(\psibf)) \, \vbf^T,
\end{align}
where $\Tcal_{\lambda, \nu}$ is a poitwise $\nu$-shrinkage operator defined as
\begin{equation}
\label{Eq:GeneralizedShrinkage}
\Tcal_{\lambda, \nu}(x) \defn \max(0, |x|-\lambda|x|^{\nu-1})\frac{x}{|x|}.
\end{equation}
For $\nu = 1$, $\nu$-shrinkage~\eqref{Eq:GeneralizedShrinkage} is
equivalent to conventional soft thresholding (See illustration in
Fig.~\ref{Fig:Shrink}). When $\nu \rightarrow 0$, it approaches hard
thresholding, which is similar to principal component analysis (PCA)
in the sense that it retains the significant few principal components.

Thus, the regularizer~\eqref{Eq:GeneralHuberRegularizer} is a
computationally tractable alternative to the rank penalty. While the
regularizer is not convex, it can still be efficiently optimized due
to closed form of its proximal operator~\eqref{Eq:HuberProximal}.
Note that due to nonconvexity of our regularizer for $\eta < 1$,
it is difficult to theoretically guarantee global convergence. However,
we have empirically observed that our algorithms converge reliably
over a broad spectrum of examples presented in Section~\ref{Sec:Experiments}. 

\subsection{Iterative optimization}
\label{Sec:IterativeOptimization}

To solve the optimization problem~\eqref{Eq:OptimizationProblem} under
the rank regularizer~\eqref{Eq:GeneralHuberRegularizer}, we first
simplify notation by defining an operator $\Bbf \defn (\Bbf_1, \dots,
\Bbf_P)$ and a vector ${\betabf \defn \Bbf\phibf = (\betabf_1, \dots,
\betabf_P)}$.

The minimization is performed using an augmented-Lagrangian (AL)
method~\cite{Nocedal.Wright2006}. Specifically, we seek the critical
points of the following AL
\begin{subequations}
\begin{align}
\Lcal(\phibf, \betabf, \sbf)  &\defn \frac{1}{2}\|\psibf - \Hbf\phibf\|_{\ell_2}^2 + \sum_{p = 1}^P \Rcal(\betabf_p) \\
&+ \frac{\rho}{2}\|\betabf - \Bbf\phibf\|_{\ell_2}^2 + \sbf^T(\betabf - \Bbf\phibf) \nonumber\\
&\hspace{-1cm}= \frac{1}{2}\|\psibf - \Hbf\phibf\|_{\ell_2}^2 + \sum_{p = 1}^P \Rcal(\betabf_p) \\
&+\frac{\rho}{2}\|\betabf - \Bbf\phibf + \sbf/\rho\|_{\ell_2}^2 - \frac{1}{2\rho}\|\sbf\|_{\ell_2}^2, \nonumber
\end{align}
\end{subequations}
where $\rho > 0$ is the quadratic parameter and $\sbf$ is the dual
variable that imposes the constraint $\betabf =
\Bbf\phibf$. Traditionally, an AL scheme
solves~\eqref{Eq:OptimizationProblem} by alternating between a joint
minimization step and an update step as
\begin{subequations}
\begin{align}
\label{Eq:AL1}
(\phibf^k, \betabf^k) &\leftarrow \argmin_{\phibf \in \R^{NT}, \betabf \in \R^{P \times B \times L}} \left\{\Lcal(\phibf, \betabf, \sbf^{k-1})\right\} \\
\sbf^k &\leftarrow \sbf^{k-1} + \rho(\betabf^k - \Bbf\phibf^k).
\end{align}
\end{subequations}
However, the joint minimization step~\eqref{Eq:AL1} is typically
computationally intensive. To reduce complexity, we
separate~\eqref{Eq:AL1} into a succession of simpler steps using the
well-established by now alternating direction method of multipliers
(ADMM)~\cite{Boyd.etal2011}. The steps are as follows
\begin{subequations}
\label{Eq:ADMM}
\begin{align}
\label{Eq:ADMM1}
\phibf^k &\leftarrow \argmin_{\phibf \in \R^{NT}} \left\{\Lcal(\phibf, \betabf^{k-1}, \sbf^{k-1})\right\} \\
\label{Eq:ADMM2}
\betabf^k &\leftarrow \argmin_{\betabf \in \R^{P \times B \times L}} \left\{\Lcal(\phibf^k, \betabf, \sbf^{k-1})\right\} \\
\label{Eq:ADMM3}
\sbf^k &\leftarrow \sbf^{k-1} + \rho(\betabf^k - \Bbf\phibf^k).
\end{align}
\end{subequations}
By ignoring the terms that do not depend on $\phibf$,~\eqref{Eq:ADMM1}
amounts to solving a quadratic problem
\begin{align}
\phibf^k &\leftarrow \argmin_{\phibf \in \R^{NT}} \left\{\frac{1}{2}\|\psibf - \Hbf\phibf\|_{\ell_2}^2 + \frac{\rho}{2}\|\Bbf\phibf - \zbf^{k-1}\|_{\ell_2}^2\right\} \nonumber\\
&\leftarrow (\Hbf^T\Hbf + \rho \Bbf^T\Bbf)^{-1}(\Hbf\psibf + \rho\Bbf^T\zbf^{k-1}), \label{Eq:QuadSubProb}
\end{align}
where $\zbf^{k-1} \defn \betabf^{k-1}+\sbf^{k-1}/\rho$. Solving this
quadratic equation is efficient since the inversion is performed on a
diagonal matrix. Similarly,~\eqref{Eq:ADMM2} is solved by
\begin{equation}
\betabf^k \leftarrow \argmin_{\betabf \in \R^{P \times B \times L}} \left\{\frac{\rho}{2}\|\betabf - \ybf^k\|_{\ell_2}^2 + \sum_{p = 1}^P \Rcal(\betabf_p)\right\},
\end{equation}
with $\ybf^k \defn \Bbf\phibf^k - \sbf^{k-1}/\rho$. This step can be
solved via block-wise application of the proximal
operator~\eqref{Eq:HuberProximal} as
\begin{equation}
\betabf_p^k \leftarrow \prox_{(\lambda/\rho)\Gcal_{\lambda, \nu}}(\Bbf_p \phibf^k - \sbf_p^{k-1}/\rho),
\end{equation}
for all $p \in [1, \dots, P]$.

\subsection{Simplified algorithm}
\label{Sec:SimplifiedAlgorithm}

The algorithm in Section~\ref{Sec:IterativeOptimization} can be
significantly simplified by decoupling the enforcement of the
data-fidelity from the enforcement of the rank-based
regularization. The simplified algorithm reduces computational
complexity while making estimation more uniform across the whole
space-time depth image.

In particular, Danielyan~\cite{Danielyan2013} has argued that, due to
inhomogeneous distribution of pixel references generated by matching
across the image, using a penalty with a single regularization
parameter highly penalizes pixels with a large number of
references. The resulting nonuniform regularization makes the
algorithm potentially oversensitive to the choice of the parameter
$\lambda$. Instead, we rely on the simplified algorithm
\begin{subequations}
\label{Eq:Dec}
\begin{align}
\label{Eq:Dec1}
\betabf_p^k &\leftarrow \argmin_{\betabf_p \in \R^{B \times L}}\left\{\frac{1}{2}\|\betabf_p - \Bbf_p \phibf^{k-1}\|_F^2 + \Rcal(\betabf_p)\right\} \\
\label{Eq:Dec2}
\phibf^k &\leftarrow \argmin_{\phibf \in \R^{NT}} \left\{\frac{1}{2}\|\psibf - \Hbf\phibf\|_{\ell_2}^2 + \frac{\rho}{2}\|\phibf - \phibftilde^k\|_{\ell_2}^2\right\}
\end{align}
\end{subequations}
where $\phibftilde^k \defn \Rbf^{-1}\Bbf^T\betabf^k$, and $\lambda > 0$ is the regularization and $\rho > 0$ is the quadratic parameters. 

To solve~\eqref{Eq:Dec1} we apply the proximal operator
\begin{equation}
\betabf_p^k \leftarrow \prox_{\lambda \Gcal_{\lambda, \nu}}(\Bbf_p\phibf^{k-1}),
\end{equation}
for all $p \in [1, \dots, P]$. Next,~\eqref{Eq:Dec2} reduces to a
linear step
\begin{equation}
\phibf^k \leftarrow (\Hbf^T \Hbf + \rho \Ibf)^{-1}(\Hbf^T\psibf + \rho \phibftilde^k).
\end{equation}
There are substantial similarities between algorithms~\eqref{Eq:ADMM}
and~\eqref{Eq:Dec}. The main differences are that we eliminated the
dual variable $\sbf$ and simplified the quadratic
subproblem~\eqref{Eq:QuadSubProb}.


\section{Experimens}
\label{Sec:Experiments}


\begin{table*}
\footnotesize
\begin{center}
\begin{tabular}{| l | c c c c | c c c c | c c c c |}
\hline
& \multicolumn{4}{c |}{Flower} & \multicolumn{4}{c |}{Lawn} & \multicolumn{4}{c |}{Road}\\
\cline{2-13}
& $2 \times$ & $3 \times$ & $4 \times$ & $5 \times$ & $2 \times$ & $3 \times$ & $4 \times$ & $5 \times$ & $2 \times$ & $3 \times$ & $4 \times$ & $5 \times$ \\
\hline
Linear & $25.62$ & $22.81$ & $21.07$ & $20.15$ & $28.32$ & $25.89$ & $24.32$ & $23.05$ & $25.44$ & $22.78$ & $21.16$ & $20.18$ \\
TV-2D & $26.52$ & $23.17$ & $21.30$ & $20.32$ & $29.97$ & $26.56$ & $24.70$ & $23.31$ & $26.30$ & $23.21$ & $21.44$ & $20.38$ \\
WTGV-2D & $26.73$ & $23.54$ & $21.68$ & $20.66$ & $30.16$ & $26.87$ & $24.94$ & $23.66$ & $26.44$ & $23.20$ & $21.38$ & $20.57$ \\
WTV-3D & $26.84$ & $23.56$ & $21.69$ & $20.72$ & $30.45$ & $27.00$ & $25.09$ & $23.68$ & $26.54$ & $23.49$ & $21.69$ & $20.73$ \\
GDS-2D & $27.76$ & $23.91$ & $21.78$ & $20.58$ & $31.27$ & $27.58$ & $25.36$ & $23.88$ & $27.39$ & $23.89$ & $21.87$ & $20.70$ \\
DS-3D & $28.00$ & $23.82$ & $21.79$ & $20.64$ & $31.37$ & $27.34$ & $25.23$ & $23.69$ & $27.30$ & $23.92$ & $21.75$ & $20.56$ \\
ADMM-3D & $29.76$ & $25.07$ & $22.58$ & $21.26$ & $33.06$ & $\bm{28.62}$ & $\bm{26.07}$ & $\bm{24.39}$ & $28.58$ & $25.18$ & $22.74$ & $21.39$ \\
GDS-3D & $\bm{30.04}$ & $\bm{25.34}$ & $\bm{22.79}$ & $\bm{21.42}$ & $\bm{32.54}$ & $28.51$ & $26.02$ & $24.36$ & $\bm{29.10}$ & $\bm{25.52}$ & $\bm{22.96}$ & $\bm{21.66}$ \\
\hline
\end{tabular}
\end{center}
{\caption{Quantitative comparison on three video sequences with added
    noise of $30$ dB. The quality of final depth is evaluated in terms
    of SNR for four different downsizing factors of $2$, $3$, $4$, and
    $5$. The best result for each scenario is
    highlighted.} \label{Tab:Table}}
\end{table*}



\begin{figure}[t]
\begin{center}
\includegraphics[width=8cm]{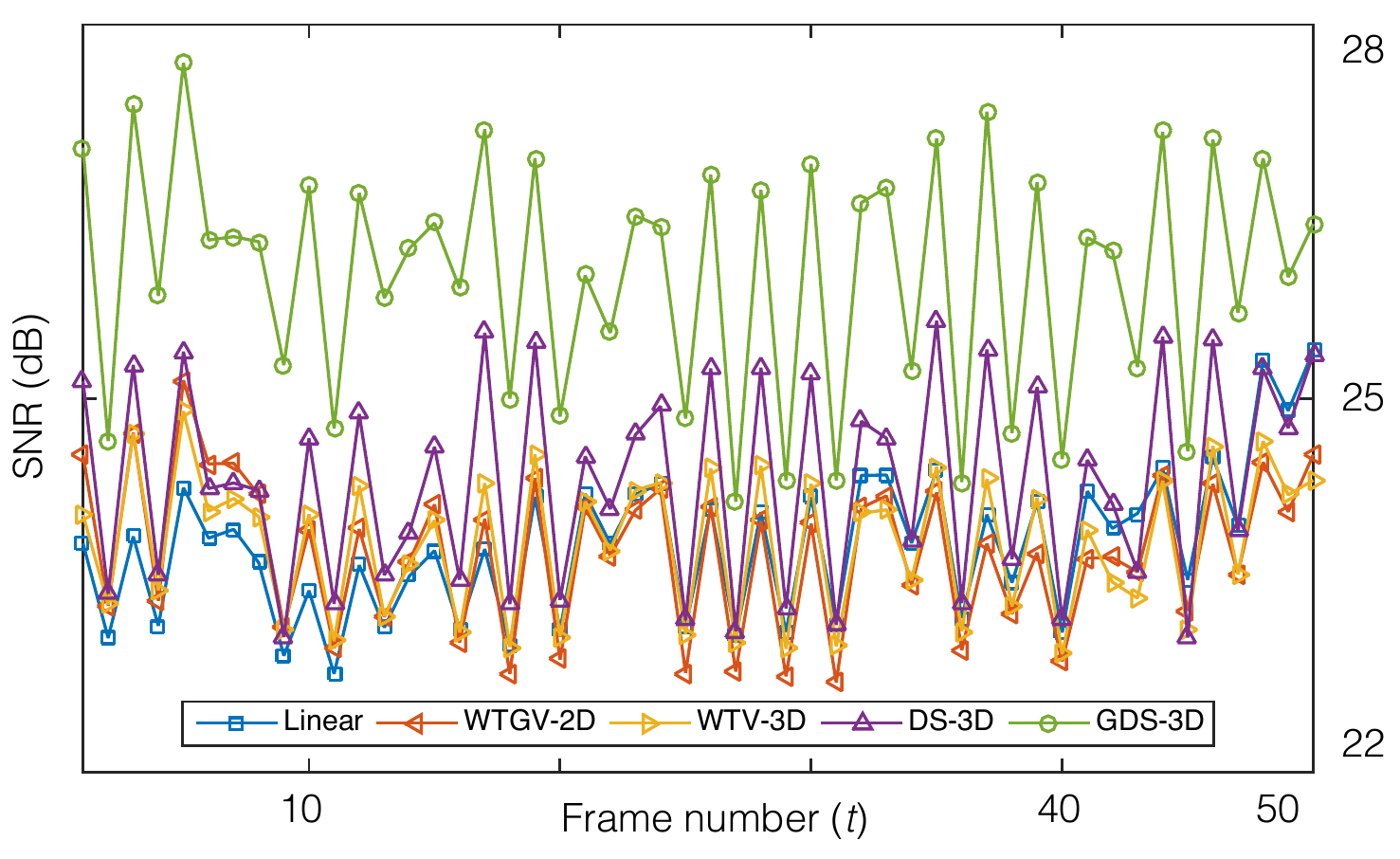}
\end{center}
\caption{Quantitative evaluation on \textit{Road} video
  sequence. Estimation of depth from its $3\times$ downsized version
  at $30$ dB input SNR. We plot the reconstruction SNR against the
  video frame number. The plot illustrates potential gains that can be
  obtained by fusing intensity and depth information in a
  motion-adaptive way.}
\label{Fig:Result2_plot}
\end{figure}

\begin{figure}[t]
\begin{center}
\includegraphics[width=8cm]{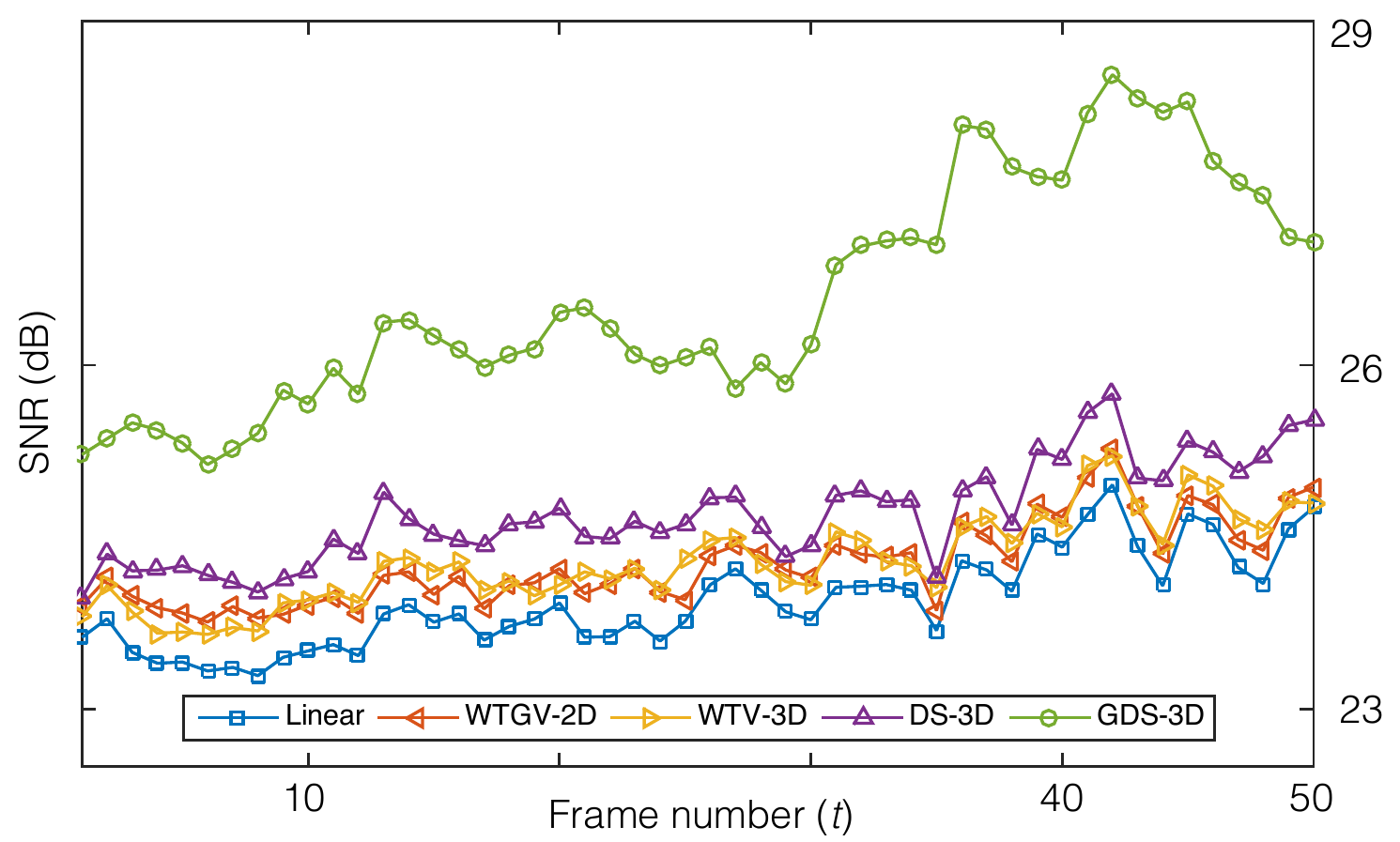}
\end{center}
\caption{Quantitative evaluation on \textit{Flower} video
  sequence. Estimation of depth from its $3\times$ downsized version
  at $30$ dB input SNR. We plot the reconstruction SNR against the
  video frame number. The plot illustrates potential gains that can be
  obtained by fusing intensity and depth information in a
  motion-adaptive way.}
\label{Fig:Result1_plot}
\end{figure}

\begin{figure*}[t]
\begin{center}
\includegraphics[width=16cm]{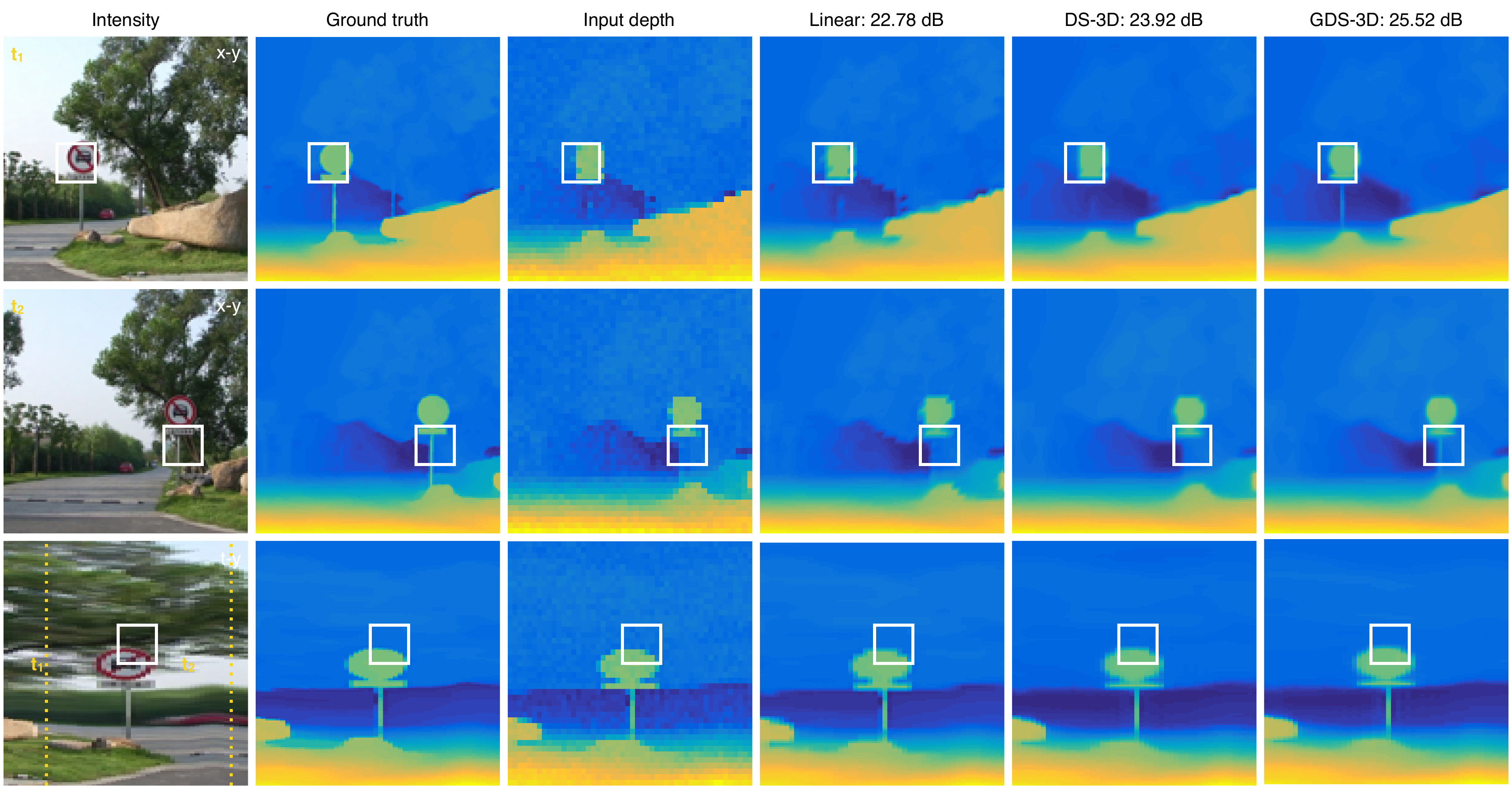}
\end{center}
\caption{Visual evaluation on \textit{Road} video sequence. Estimation of depth from its $3 \times$ downsized version at $30$ dB input SNR. Row 1 shows the data at time instance $t = 9$. Row 2 shows the data at the time instance $t = 47$. Row 3 shows the $t$-$y$ profile of the data at $x = 64$. Highlights indicate some of the areas where depth estimated by GDS-3D recovers details missing in the depth estimate of DS-3D that does not use intensity information.}
\label{Fig:Result2_img}
\end{figure*}

\begin{figure*}[t]
\begin{center}
\includegraphics[width=16cm]{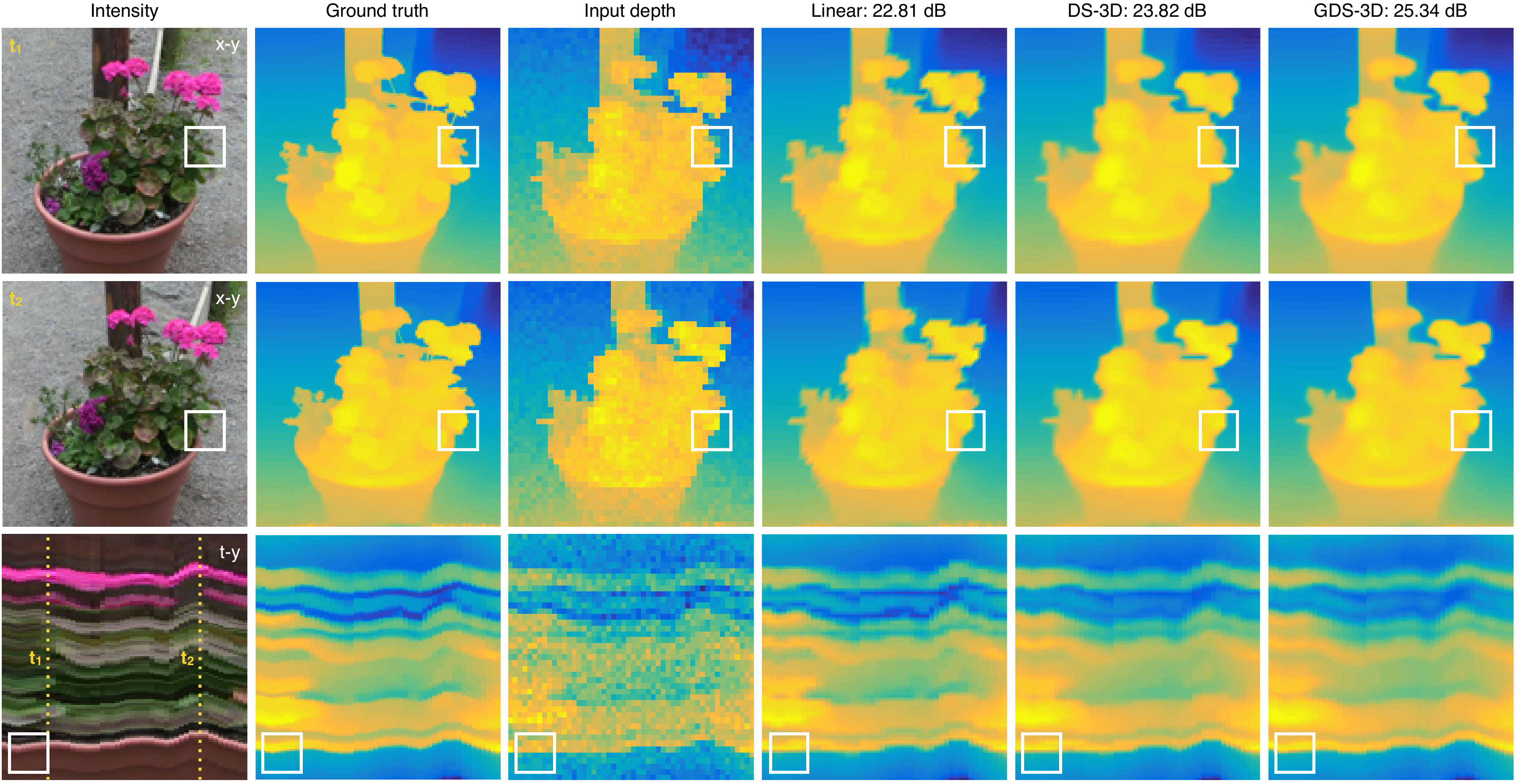}
\end{center}
\caption{Visual evaluation on \textit{Flower} video
  sequence. Estimation of depth from its $3 \times$ downsized version
  at $30$ dB input SNR. Row 1 shows the data at time instance $t =
  10$. Row 2 shows the data at the time instance $t = 40$. Row 3 shows
  the $t$-$y$ profile of the data at $x = 64$. Highlights indicate
  some of the areas where depth estimated by GDS-3D recovers details
  missing in the depth estimate of DS-3D that does not use intensity
  information.}
\label{Fig:Result1_img}
\end{figure*}

\begin{figure*}[t]
\begin{center}
\includegraphics[width=16cm]{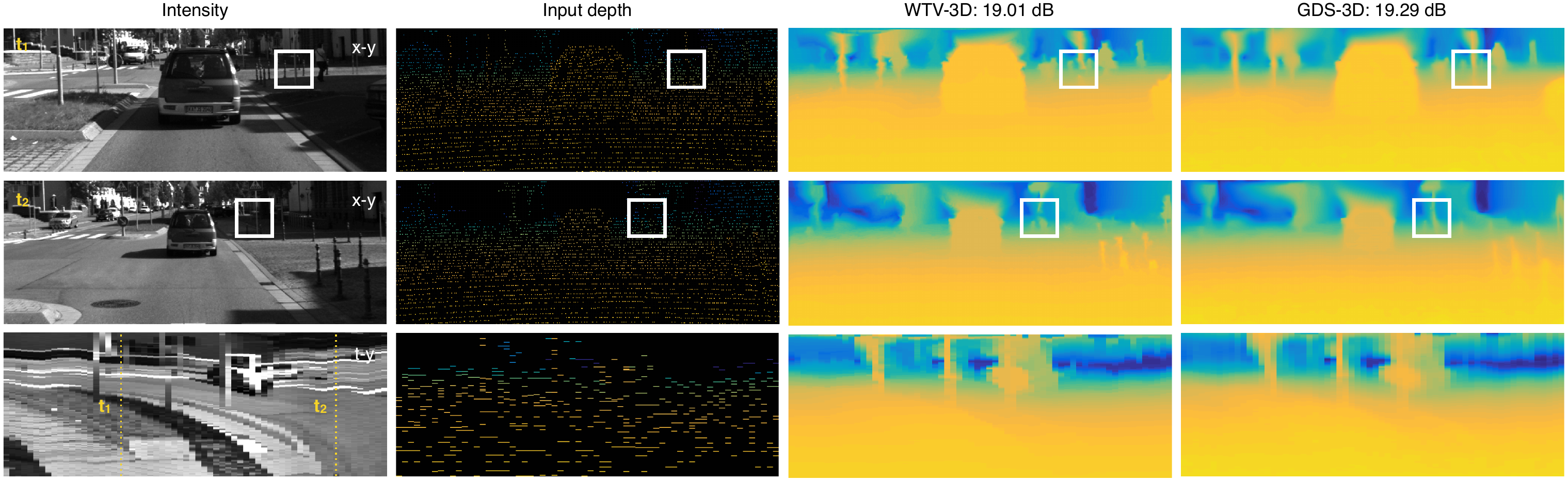}
\end{center}
\caption{Visual evaluation on \textit{KITTI} dataset. Estimation of depth images of size $192 \times 512$ with $64$ time frames from $247794$ lidar measurements, which corresponds to a measurement rate of just $3.94\%$. Row 1 shows the data at time instance $t = 20$. Row 2 shows the data at the time instance $t = 53$. Row 3 shows the $t$-$y$ profile of the data at $x = 96$. Highlights indicate some of the areas where depth estimated by GDS-3D recovers details missing in the WTV-3D estimate. }
\label{Fig:Result3_img}
\end{figure*}


To verify our development, we report results on extensive
simulations using our guided depth superresolution algorithms. In
particular, we compare results of both the ADMM approach (denoted
\textit{ADMM-3D}) and its simplified variant (denoted \textit{GDS-3D})
against six alternative methods.

As the first and the simplest reference method, we consider standard
linear interpolation (\textit{Linear}). In addition to linear
interpolation, we consider methods relying in some form of total
variation (TV) regularization, one of the most widely used
regularizers in the context of image reconstruction due to its ability
to reduce noise while preserving image
edges~\cite{Rudin.etal1992}. Specifically, we consider depth
interpolation using TV-regularized least squares on a frame-by-frame
basis (\textit{TV-2D}). We also consider the weighted-TV formulation
proposed by Ferstl \etal~\cite{Ferstl.etal2013} (\textit{WTGV-2D}),
also operating on a frame-by-frame basis. This formulation uses a
weighted anisotropic total generalized variation, where weighting is
computed using the guiding intensity image, thus promoting edge
co-occurrence in both modalities. Finally, we consider a weighted-TV
formulation which includes time, i.e., multiple frames
(\textit{WTV-3D}), with weights computed using the guiding intensity
image, as before.

We also compare these methods to two variations of our algorithm. To
illustrate the potential gains of our method due to temporal
information, we run our simplified algorithm on a frame-by-frame basis
(\textit{GDS-2D}), i.e., using no temporal information. Similarly, to
illustrate the gains due to intensity information, we also run the
simplified algorithm by performing block matching on the
low-resolution depth as opposed to intensity (\textit{DS-3D}). This
last approach conceptually corresponds to denoising the initial depth
using our motion adaptive low-rank prior.

To improve convergence speed, we initialized all the iterative
algorithms with the solution of linear interpolation. The
regularization parameters of TV-2D, WTGV-2D, and WTV-3D were optimized
for the best signal-to-noise ratio (SNR) performance. Similarly, the
regularization parameters $\lambda$ for GDS-2D, DS-3D, ADMM-3D, and
GDS-3D were optimized for the best SNR performance. However, in order
to reduce the computational time, the selection was done from a
restricted set of three predetermined parameters.  The methods TV-2D,
WTV-3D, GDS-2D, DS-3D, ADMM-3D, and GDS-3D were all run for a maximum
of $t_{\text{\tiny max}} = 100$ iterations with an additional stopping
criterion based on measuring the relative change of the solution in
two successive iterations
\begin{equation}
\frac{\|\phibf^k - \phibf^{k-1}\|_{\ell_2}}{\|\phibf^{k-1}\|_{\ell_2}}
\leq \delta,
\end{equation}
where $\delta = 10^{-4}$ is the desired tolerance level. We selected
other parameters of WTGV-2D as suggested in the code provided by the
authors; in particular, we run the algorithm for a maximum of $1000$
iterations with the stopping tolerance of $0.1$. In all experiments,
the patch size was set to $5 \times 5$ pixels, the space-time window
size to $11 \times 11 \times 3$ pixels, and the number of similar
patches was fixed to $10$. Parameters $\nu$ and $\rho$ were hand
selected to $0.02$ and $1$, respectively.

For quantitative comparison of the algorithms, we rely on the data-set
by Zhang~\etal~\cite{Zhang.etal2009}, which consists of three video
sequences \textit{Flower}, \textit{Lawn}, and \textit{Road},
containing both intensity and depth information on the scenes. We
considered images of size $128 \times 128$ with $50$ time frames. The
ground truth depth was downsized by factors of $2$, $3$, $4$, and $5$,
and was corrupted by additive Gaussian noise corresponding to SNR of
$30$ dB. Table~\ref{Tab:Table} reports the SNR of superresolved depth
for all the algorithms and downsizing
factors. Figures~\ref{Fig:Result2_plot} and~\ref{Fig:Result1_plot}
illustrate the evolution of SNR against the frame number for
\textit{Road} and \textit{Flower}, respectively, at downsizing factor
of $3$. The effectiveness of the proposed approach can also be
appreciated visually in Figs.~\ref{Fig:Result2_img}
and~\ref{Fig:Result1_img}.

For additional validation, we test the algorithm on the KITTI
dataset~\cite{Geiger.etal2013}. The dataset contains intensity images
from PointGray Flea2 cameras and 3D point clouds from a Velodyne
HDL-64E, which have been calibrated a priori using specific known
targets. We consider a region of interest of $192 \times 512$ pixels
with $64$ time frames. The $495588$ total lidar measurements are
randomly and uniformly split into a reconstruction and a validation
sets. Note that this implies a depth measurement rate of just
$3.94\%$. We then estimate depth from the reconstruction set using
WTV-3D, as well as GDS-3D, and use the validation set to evaluate the
quality of the results; these are illustrated in
Fig.~\ref{Fig:Result3_img}.

The examples and simulations in this section, validate our claim: the
quality of estimated depth can be considerably boosted by properly
incorporating temporal information into the reconstruction
procedure. Comparison of GDS-2D against GDS-3D highlights the
importance of additional temporal information. The approach we propose
is implicitly motion adaptive and can thus preserve temporal edges
substantially better than alternative approaches such as
WTV-3D. Moreover, comparison between DS-3D and GDS-3D highlights that
the usage of intensity significantly improves the performance of the
algorithm. Note also the slight improvement in SNR of GDS-3D over
ADMM-3D. This is consistent with the arguments in~\cite{Danielyan2013}
that suggest to decouple data-fidelity from the enforcement of prior
constraints when using block-matching-based methods.


\section{Conclusion}
\label{Sec:Conclusion}
We presented a novel motion-adaptive approach for guided
superresolution of depth maps. Our method identifies and groups similar
patches from several frames, which are then supperresolved using a rank regularizer.
Using this approach, we can produce high-resolution 
depth sequences from significantly down-sized low-resolution ones. 
Compared to the standard techniques, the proposed method 
preserves temporal edges in the solution and effectively 
mitigates noise in practical configurations.

While our formulation has higher computational complexity than
standard approaches that process each frame individually, it
allows us to incorporate a very effective regularization for stabilizing the inverse problem associated with superresolution. The algorithms we describe enable efficient computation and straightforward implementation by reducing the problem to a succession of straightforward operations. 
Our experimental results demonstrate the considerable 
benefits of incorporating time and motion adaptivity into
inverse-problems for depth estimation.


\bibliographystyle{IEEEtran}

\begin{thebibliography}{10}
\providecommand{\url}[1]{#1}
\csname url@samestyle\endcsname
\providecommand{\newblock}{\relax}
\providecommand{\bibinfo}[2]{#2}
\providecommand{\BIBentrySTDinterwordspacing}{\spaceskip=0pt\relax}
\providecommand{\BIBentryALTinterwordstretchfactor}{4}
\providecommand{\BIBentryALTinterwordspacing}{\spaceskip=\fontdimen2\font plus
\BIBentryALTinterwordstretchfactor\fontdimen3\font minus
  \fontdimen4\font\relax}
\providecommand{\BIBforeignlanguage}[2]{{%
\expandafter\ifx\csname l@#1\endcsname\relax
\typeout{** WARNING: IEEEtran.bst: No hyphenation pattern has been}%
\typeout{** loaded for the language `#1'. Using the pattern for}%
\typeout{** the default language instead.}%
\else
\language=\csname l@#1\endcsname
\fi
#2}}
\providecommand{\BIBdecl}{\relax}
\BIBdecl

\bibitem{Diebel.Thrun2005}
J.~Diebel and S.~Thrun, ``An application of markov random ffield to range
  sensing,'' in \emph{Proc. Advances in Neural Information Processing Systems
  18}, vol.~18, Vancouver, BC, Canada, December 5-8, 2005, pp. 291--298.

\bibitem{Kopf.etal2007}
J.~Kopf, M.~F. Cohen, D.~Lischinski, and M.~Uyttendaele, ``Joint bilateral
  upsampling,'' in \emph{ACM Trans. Graph. (Proc. {SIGGRAPH} 2007)}, vol.~26,
  San Diego, CA, USA, August 5-9, 2007, p.~96.

\bibitem{Yang.etal2007}
Q.~Yang, R.~Yang, J.~Davis, and D.~Nister, ``Spatial-depth super resolution for
  range images,'' in \emph{Proc. {IEEE} Conf. Computer Vision and Pattern
  Recognition ({CVPR})}, Minneapolis, MN, USA, June 17-22, 2007, pp. 1--8.

\bibitem{Chan.etal2008}
D.~Chan, H.~Buisman, C.~Theobalt, and S.~Thrun, ``A noise-aware filter for
  real-time depth upsampling,'' in \emph{ECCV Workshop on Multicamera and
  Multimodal Sensor Fusion Algorithms and Applications}, 2008.

\bibitem{Liu.etal2013}
M.-Y. Liu, O.~Tuzel, and Y.~Taguchi, ``Joint geodesic upsampling of depth
  images,'' in \emph{Proc. {IEEE} Conf. Computer Vision and Pattern Recognition
  ({CVPR})}, Portland, {OR}, {USA}, June 23-28, 2013, pp. 169--176.

\bibitem{Dolson.etal2010}
J.~Dolson, J.~Baek, C.~Plagemann, and S.~Thrun, ``Upsampling range data in
  dynamic environments,'' in \emph{Proc. {IEEE} Conf. Computer Vision and
  Pattern Recognition ({CVPR})}, vol.~23, San Francisco, CA, USA, June 13-18,
  2010, pp. 1141--1148.

\bibitem{He.etal2010}
K.~He, J.~Sun, and X.~Tang, ``Guided image filtering,'' in \emph{Proc. {ECCV}},
  Hersonissos, Greece, September 5-11, 2010, pp. 1--14.

\bibitem{He.etal2013}
------, ``Guided image filtering,'' \emph{IEEE Trans. Patt. Anal. and Machine
  Intell.}, vol.~35, no.~6, pp. 1397--1409, June 2013.

\bibitem{Candes.etal2006}
E.~J. Cand{\`e}s, J.~Romberg, and T.~Tao, ``Robust uncertainty principles:
  Exact signal reconstruction from highly incomplete frequency information,''
  \emph{IEEE Trans. Inf. Theory}, vol.~52, no.~2, pp. 489--509, February 2006.

\bibitem{Donoho2006}
D.~L. Donoho, ``Compressed sensing,'' \emph{IEEE Trans. Inf. Theory}, vol.~52,
  no.~4, pp. 1289--1306, April 2006.

\bibitem{Li.etal2012}
Y.~Li, T.~Xue, L.~Sun, and J.~Liu, ``Joint example-based depth map
  super-resolution,'' in \emph{Proc. IEEE Int. Con. Multi.}, Melbourne, {VIC},
  Australia, July 9-13, 2012, pp. 152--157.

\bibitem{Ferstl.etal2013}
D.~Ferstl, C.~Reinbacher, R.~Ranftl, M.~Ruether, and H.~Bischof, ``Image guided
  depth upsampling using anisotropic total generalized variation,'' in
  \emph{Proc. {IEEE} Conf. Computer Vision and Pattern Recognition ({CVPR})},
  Sydney, {NSW}, Australia, December 1-8, 2013, pp. 993--1000.

\bibitem{Gong.etal2014}
X.~Gong, J.~Ren, B.~Lai, C.~Yan, and H.~Qian, ``Guided depth upsampling via a
  cosparse analysis model,'' in \emph{Proc. {IEEE} {CVPR} {WKSHP}}, Columbus,
  {OH}, {USA}, June 23-28, 2014, pp. 738--745.

\bibitem{Huang.etal2015}
W.~Huang, X.~Gong, and M.~Y. Yang, ``Joint object segmentation and depth
  upsampling,'' \emph{IEEE Signal Process. Lett.}, vol.~22, no.~2, pp.
  192--196, February 2015.

\bibitem{Schouon.etal2009}
S.~Schouon, C.~Theobalt, J.~Davis, and S.~Thrun, ``{L}idar{B}oost: {D}epth
  superresolution for {ToF} {3D} shape scanning,'' in \emph{Proc. {IEEE} Conf.
  Computer Vision and Pattern Recognition ({CVPR})}, Miami, {FL}, {USA}, June
  20-25, 2009, pp. 343--350.

\bibitem{Buades.etal2010}
A.~Buades, B.~Coll, and J.~M. Morel, ``Image denoising methods. {A} new
  nonlocal principle.'' \emph{SIAM Rev}, vol.~52, no.~1, pp. 113--147, 2010.

\bibitem{Dabov.etal2007}
K.~Dabov, A.~Foi, V.~Katkovnik, and K.~Egiazarian, ``Image denoising by sparse
  {3-D} transform-domain collaborative filtering,'' \emph{IEEE Trans. Image
  Process.}, vol.~16, no.~16, pp. 2080--2095, August 2007.

\bibitem{Yang.Jacob2013}
Z.~Yang and M.~Jacob, ``Nonlocal regularization of inverse problems: A unified
  variational framework,'' \emph{IEEE Trans. Image Process.}, vol.~22, no.~8,
  pp. 3192--3203, August 2013.

\bibitem{Danielyan.etal2012}
A.~Danielyan, V.~Katkovnik, and K.~Egiazarian, ``{BM3D} frames and variational
  image deblurring,'' \emph{IEEE Trans. Image Process.}, vol.~21, no.~4, pp.
  1715--1728, April 2012.

\bibitem{Huhle.etal2010}
B.~Huhle, T.~Schairer, P.~Jenke, and W.~Strasser, ``Fusion of range and color
  images for denoising and resolution enhancement with a non-local filter,''
  \emph{Comput. Vis. Image Underst.}, vol. 114, no.~12, pp. 1336--1345,
  December 2010.

\bibitem{Park.etal2011}
J.~Park, H.~Kim, Y.-W. Tai, M.~S. Brown, and I.~Kweon, ``High quality depth map
  upsampling for {3D}-{TOF} cameras,'' in \emph{Proc. {IEEE} Int. Conf. Comp.
  Vis.}, Barcelona, Spain, November 6-13, 2011, pp. 1623--1630.

\bibitem{Lu.etal2014}
S.~Lu, X.~Ren, and F.~Liu, ``Depth enhancement via low-rank matrix
  completion,'' in \emph{Proc. {IEEE} Conf. Computer Vision and Pattern
  Recognition ({CVPR})}, Columbus, {OH}, {USA}, June 23-28, 2014, pp.
  3390--3397.

\bibitem{fazel2002matrix}
M.~Fazel, ``Matrix rank minimization with applications,'' Ph.D. dissertation,
  PhD thesis, Stanford University, 2002.

\bibitem{Chartrand2012}
R.~Chartrand, ``Nonconvex splitting for regularized {Low-Rank + Sparse}
  decomposition,'' \emph{IEEE Trans. Signal Process.}, vol.~60, no.~11, pp.
  5810--5819, November 2012.

\bibitem{Hu.etal2012}
Y.~Hu, S.~G. Lingala, and M.~Jacob, ``A fast majorize-minimize algorithm for
  the recovery of sparse and low-rank matrices,'' \emph{IEEE Trans. Image
  Process.}, vol.~21, no.~2, pp. 742--753, February 2012.

\bibitem{Yoon.etal2014}
H.~Yoon, K.~S. Kim, D.~Kim, Y.~Bresler, and J.~C. Ye, ``Motion adaptive
  patch-based low-rank approach for compressed sensing cardiac cine {MRI},''
  \emph{IEEE Trans. Med. Imag.}, vol.~33, no.~11, pp. 2069--2085, November
  2014.

\bibitem{Nocedal.Wright2006}
J.~Nocedal and S.~J. Wright, \emph{Numerical Optimization}, 2nd~ed.\hskip 1em
  plus 0.5em minus 0.4em\relax Springer, 2006.

\bibitem{Boyd.etal2011}
S.~Boyd, N.~Parikh, E.~Chu, B.~Peleato, and J.~Eckstein, ``Distributed
  optimization and statistical learning via the alternating direction method of
  multipliers,'' \emph{Foundations and Trends in Machine Learning}, vol.~3,
  no.~1, pp. 1--122, 2011.

\bibitem{Danielyan2013}
A.~Danielyan, ``Block-based collaborative 3-{D} transform domain modeing in
  inverse imaging,'' Publication 1145, {T}ampere {U}niversity of {T}echnology,
  May 2013.

\bibitem{Rudin.etal1992}
L.~I. Rudin, S.~Osher, and E.~Fatemi, ``Nonlinear total variation based noise
  removal algorithms,'' \emph{Physica D}, vol.~60, no. 1--4, pp. 259--268,
  November 1992.

\bibitem{Zhang.etal2009}
L.~Zhang, S.~Vaddadi, H.~Jin, and S.~K. Nayar, ``Multiple view image
  denoising,'' in \emph{Proc. {IEEE} Conf. Computer Vision and Pattern
  Recognition ({CVPR})}, Miami, FL, USA, June 20-25, 2009, pp. 1542--1549.

\bibitem{Geiger.etal2013}
A.~Geiger, P.~Lenz, C.~Stiller, and R.~Urtasun, ``Vision meets robotics: The
  {KITTI} dataset,'' \emph{International Journal of Robotics Research},
  vol.~32, no.~11, pp. 1231--1237, November 2013.

\end{thebibliography}


\end{document}